\documentclass[acmsmall,screen]{acmart}

\setcopyright{none}
\renewcommand\footnotetextcopyrightpermission[1]{}

\begin{document}

\title{When Architectural Spaces Start to Act}

\author{Alex Binh Vinh Duc Nguyen}
\email{alexbvd.nguyen@uantwerpen.be}
\orcid{0000-0001-5026-474X}
\affiliation{
  \institution{Faculty of Design Sciences, University of Antwerp}
  \city{Antwerp}
  \country{Belgium}
}


\begin{abstract}
Architectural spaces are increasingly gaining robotic capabilities. Architectural robots such as robotic partitions, ceilings, furniture, façades, and other spatial components can autonomously move and reconfigure in response to occupants' needs and activities. Yet, when architectural robots start shaping the spatial conditions in which people continuously live, work, and interact, we may begin to experience the architectural space itself as robotic. What does this shift mean for how we inhabit such spaces? Drawing on our previous studies, I reflect on how architectural robots interact with people not only through their motions, but also through the spatial configurations that persist after those motions stop; how their actions may benefit some occupants while disadvantaging others; and how multiple robots may together make a whole architectural space appear to act. I argue that as computation gains the capacity to physically reconfigure our environments, interaction extends beyond the moment of computational action into the spatial conditions that people continue to inhabit. Understanding robotic architecture therefore requires us to look beyond short-term human-robot interaction and toward how people inhabit, negotiate, and maintain control over spaces that can autonomously reconfigure.
\end{abstract}



\maketitle

\fancypagestyle{standardpagestyle}{
  \fancyhf{}
  \renewcommand{\headrulewidth}{0pt}
  \renewcommand{\footrulewidth}{0pt}
}

\fancypagestyle{firstpagestyle}{
  \fancyhf{}
  \renewcommand{\headrulewidth}{0pt}
  \renewcommand{\footrulewidth}{0pt}
}

\pagestyle{standardpagestyle}
\thispagestyle{firstpagestyle}

\section{Robotic architecture: when space is no longer the background}

Imagine an architectural space that is dear to you. Perhaps you would think of your childhood bedroom, a window that you always look out from, or a dining room where your family shares everyday meals. Perhaps you would remember the layout of the whole space, or only a corner of it: the shape of a piece of furniture, the surface of a wall, the smell of wood. Perhaps you are in that architectural space right now, and this paragraph makes you start looking around and notice everything that it has to offer. For much of our daily life, we rarely pay continuous attention to the architectural spaces we inhabit. Once a space becomes familiar, it recedes into the background while we work, eat, sleep, talk, and live within it.

Although architectural spaces mostly exist in the background of our attention, we continuously interact with them through \textit{adaptations}. We consciously and subconsciously adapt our everyday activities, sometimes even our preferences, to fit ourselves into these architectural spaces. In turn, we also adapt these spaces to accommodate our needs, sometimes by physically reconfiguring them. We open a window when a room becomes stuffy, close a curtain when sunlight becomes uncomfortable, move a chair closer to a table, or rearrange furniture when our activities change. Spatial reconfigurations may differ greatly in scale: sometimes they take place through minor adjustments, and sometimes through major renovation projects. Yet, in our everyday lives, spatial reconfigurations usually remain within our control. We decide what should be reconfigured, we perform the reconfigurations ourselves, or at least we understand why a reconfiguration is happening.

Now, I ask you to remember again the same space that is dear to you. Imagine the window that you used to look out from; now it can automatically open or close to regulate the air quality of the room. Imagine the dinner table with your family, now with an additional capability to move and reconfigure its shape to act both as a kitchen island during cooking and a dining table during meals. Imagine the chair that you used to sit on; now it can move across the room to help your grandma when she has difficulties walking. Here, the architecture itself can initiate a physical spatial reconfiguration. Perhaps such robotic architecture sounds convenient, perhaps unsettling. Either way, it remains difficult to imagine how we would inhabit it in the same way that we currently inhabit our spaces.

I would like to emphasise that this idea is not new. Architects, designers, and roboticists have long envisioned how interactive technologies can become embedded within the physical fabric of everyday environments, ranging from responsive façades to robotic furniture and whole reconfigurable rooms~\cite{Green2016}. The field of human-building interaction (HBI) is increasingly applying the principles of human-computer interaction (HCI) toward understanding buildings as networks of interactive systems~\cite{Alavi2019}. The emerging interest in robotic furniture within the field of human-robot interaction (HRI) has shown how ordinary spatial components can perform movements with gestural qualities that people interpret as conveying cognition and intent~\cite{Hoffman2014}. These developments connect robotic architecture to a longer trajectory of ubiquitous computing, which has long envisioned computation as receding into the fabric of everyday environments~\cite{Weiser1991}, while everyday objects and architectural surfaces become interfaces between digital information and the physical world~\cite{Ishii1997}. Robotic architecture extends this trajectory further: computation can now sense our activities and physically reorganise the environment itself. What remains unclear is how we would continuously inhabit such ever-changing architectural spaces.

In this article, I thus explore this problem through a thought exercise: How would ordinary lives unfold within a robotic architectural space that is sometimes the background and sometimes is not? How do we inhabit this shelter that should provide us with safety and security while also being capable of changing itself? What changes for HCI when the output of computation is a spatial condition that people must continue to inhabit? Throughout this article, I use \textit{architectural robots} to describe individual robotic components that act as architectural elements, and \textit{robotic architecture} to describe the larger inhabited environment that emerges through their actions.

\section{Architectural robots: when interaction continues beyond motion}

It may be tempting to approach a robotic architectural space as just another kind of robot, so that knowledge long studied in HRI, including agency, intention, predictability, safety, and control, can be directly transferred to understand how people may experience such spaces. Yet, while robots mostly interact with people through motion, architecture typically interacts with people through non-motion. We understand the capabilities of robots through their actions. A building may adjust some of its components automatically, may incorporate robotic capability to reconfigure its spatial layout, adapt its façade, or even rotate the whole building to follow the sun. Yet, we experience the building mostly when it has stopped moving. We interact with its architectural spaces for a much longer period of time than we actually interact with the dynamic transitions between those spaces. A robotic action may last seconds, while the resulting spatial configuration may shape what people see, where they move, and what they can do for minutes, hours, or even longer.

We encountered these different ways of interaction while studying architectural robots: autonomous mobile robots that embody architectural elements. In one of our studies, we invited experts from architecture, performance, visual media, and robotics to design how a mobile robotic partition might autonomously convey architectural intentions to an occupant, such as shielding an occupant from a disturbance, showing a direction, or asking them to relocate~\cite{Nguyen2025}. While our experts employed a diverse range of motion-based gestures, ranging from familiar emblematic (e.g. nodding, shaking the head) to deictic (e.g. pointing) motion patterns, they also designed location-based gestures that relied heavily on the spatial consequences of the partition once it stopped moving. By reconfiguring the surrounding space, the partition could reveal a passage, regulate visual access, communicate a direction, or even purposefully create discomfort for occupants. We termed these \textit{architectonic gestures}, namely motions whose communicative meanings emerge from the spatial configurations they produce. We argue that while such gestures may also occur with other robots, they are particularly relevant to architectural and furniture robots, whose embodiments can directly reconfigure the surrounding space.

Our studies with the robotic partition also revealed that understanding how spatial configurations affect people's experience of robotic architecture is particularly complex in shared environments~\cite{Balci2025}. Within the field of HBI, it is well established that people can experience the same space differently, and a space that is comfortable for one occupant may not adequately support another. Consequently, a spatial configuration resulting from an action of an architectural robot may inadvertently prioritise one occupant over another. Imagine a shared, open office environment where multiple spatial components can robotically reconfigure. In this space, closing a façade panel may reduce glare for one workstation while limiting views for another. Rearranging a group of tables to support collaboration among one group of workers may disturb others doing individual focused work. Repeatedly being disadvantaged by such configurations may not only create discomfort for an occupant, but also undermine their trust in the architectural robots and willingness to work in such an office. Here, the consequences of human-robot interaction arise not only through dynamic motion, but also through the spatial configurations that remain after the motion stops.

There is another complication. Once an architectural robot stops moving, people may no longer recognise it as a robot at all. Because its embodiment affords spatial functionality, an architectural robot can recede into the background as an everyday object once its robotic motion ends. Our in-the-wild study with robotic public furniture demonstrated this phenomenon~\cite{Yu2026}. Through a purposefully designed gesturing sequence, a set of robotic benches was able to convey different affordances to passers-by. They transitioned between being attention-seeking \textit{robots}, to being \textit{spatial elements} that distributed the attention of passers-by into the surroundings, and finally to being \textit{benches} that passers-by assumed were part of the public infrastructure. Depending on when passers-by encountered the benches, they perceived different affordances and interacted with them accordingly. Someone encountering a static bench could assume that it was ordinary public infrastructure and become startled when it later moved. In contrast, someone encountering the same bench when it was interacting with another person would expect it to do the same when they approached.

In short, our studies show that architectural robots can interact with people by gesturing through both their motion and their spatial consequences, which together create a design challenge: an architectural robot must manage people's expectations of its robotic capability, even when it appears as an ordinary static spatial element. Once it starts moving, the architectural robot must also manage how its motion may communicate an intention to people. When the motion stops, the architectural robot must still manage how its spatial consequences may be perceived by different people sharing the same space. This phenomenon echoes the long-standing ambition of ubiquitous computing for computation to recede into everyday life~\cite{Weiser1991}, but introduces a physical complication: what has receded into the infrastructure can later move and change the environment around us.

\section{When architecture can decide: a quest for environmental control}

As humans, we seek some level of control over our surrounding environments. We prefer architectural spaces that give us agency: a house that we can lock and know that strangers cannot enter, or a seat that shelters our back while maintaining a view of our surroundings. If an architectural space can change, we expect that there would be some interface through which we can influence those changes, such as a button for us to turn the light on and off, a panel for us to adjust the heater, or a proximity zone around an automated door that we know will cause it to open when we approach. This desire for control is important for us, to the extent that reduced perceived control can contribute to stress and lack of satisfaction~\cite{Kwon2019}. This could be because the control interface is too complicated for us to understand, because the building component does not react the way that we expect, or because the presence of other people makes us feel that we should not change the space in order to avoid affecting them.

For architectural robots, maintaining perceived control for occupants involves two intertwined questions. The first concerns the robotic action: can an occupant understand why an architectural element is moving, predict its action, or stop it when necessary? The second concerns the spatial consequences that the robotic action produces: can an occupant understand, accept, and influence the resulting spatial configuration? Imagine a robotic chair moving within a room autonomously. Its trajectory might be perfectly predictable and technically safe. Yet the occupants may still deem such robotic action unacceptable because its destination blocks their circulation, provides an inappropriate place to sit, or simply because they do not recognise the need for changing the spatial configuration at that moment. If the robotic chair offers no way for occupants to deactivate, stop, or override its actions, they may perceive not only the chair but the surrounding space as uninhabitable.

In shared environments, this complexity becomes more prominent, because different occupants may perceive the same spatial configuration resulting from a robotic action in different ways depending on their current needs. An architectural robot may appear to act appropriately for one occupant because it is reacting to their needs, such as a robotic partition moving to cover the glare from a window. Yet, from the perspective of another occupant, the same robotic action may distract them from working and thus may be perceived as inappropriate. As the spatial needs of individual occupants are not always immediately perceivable to others, this can increase the risk of misinterpretation and mistrust. This problem thus raises questions of negotiation: Which occupants should receive priority when their spatial needs conflict? When should the architectural robot decide on a spatial reconfiguration, and when should it ask the occupants to negotiate? How can an occupant contest a configuration that disadvantages them? And how can the architectural robot learn from this negotiation process to act more appropriately in the future?

The design problem becomes even more interesting in spaces where multiple architectural robots must act together. Imagine a bedroom autonomously reconfiguring into a home office. The bed folds into the ceiling, a desk and chair move into working positions, and the window darkens to reduce glare. When these robotic actions occur in a well-coordinated sequence, occupants may no longer perceive them as independent robots performing separate tasks. They may experience the whole room as acting toward a consistent purpose: to support their activity. Yet, if the sequence of robotic actions is carried out imperfectly, with one architectural robot acting inconsistently with another, the occupant may become confused about their purposes or even feel unsafe being in such a room. In both cases, occupants may begin to attribute agency to the whole architectural space, which then raises the question: how should an occupant interact with such a multi-robot room so that they can still maintain sufficient perceived control? Should one of the architectural robots act as a "concierge" on behalf of the whole room, through which the occupant could interact with the other robots? Should there be a separate interface to mediate the interaction between the occupant and the architectural robots? Or, more provocatively, should there be no explicit interaction at all, and should the architecture act robotically only when the occupants are not around?

These questions of interactivity can be reframed as when and how architectural robots should transition between different affordances: as a robot, a spatial element, or part of the infrastructure. Imagine a room in which furniture periodically reconfigures simply to make its robotic capabilities visible to unfamiliar occupants. Such an environment would quickly become exhausting to inhabit. Yet, at the opposite extreme, if an architectural robot remains static for long periods and becomes indistinguishable from ordinary infrastructure, occupants may gradually stop anticipating that it could move at all. A sudden autonomous action may then become surprising, especially for occupants who have not encountered the object in motion. The question for robotic architecture, therefore, is how to balance its different affordances: making occupants aware that their architectural space may autonomously reconfigure; orchestrating the when, why, and how of such reconfigurations in an appropriate and understandable way; and allowing occupants to contest, stop, or override such reconfigurations if they desire.

\section{Research directions: where do we go from here?}

It is becoming clear that designing and understanding the interaction between humans and architectural robots is challenging. To answer our initial question of how ordinary lives would unfold within a robotic architectural space, research efforts need to move beyond spontaneous interactions between a single occupant and a single architectural robot, and investigate the longer-term inhabitation of architectural environments that are shared by multiple occupants and embody multiple robots. While such environments remain uncommon, their development raises several problems that concern the broader computing community.

First, computational representations of robotic environments need to account not only for robot trajectories, but also for their spatial consequences: who can see whom, which routes remain accessible, what activities are enabled or constrained, and how these conditions differ for different human (and non-human) occupants. Second, mechanisms for autonomous reconfiguration need to account for multiple occupants whose needs may conflict, with appropriate interfaces for shared control and negotiation. Third, interaction with robotic environments needs to remain legible even when its computational capabilities recede into the background. Addressing these problems will require combining knowledge from HBI and smart building research on occupant control, HRI research on interaction with distributed robots, and architectural design research on how spatial configurations accommodate different forms of occupancy and use. For computing research, perhaps the important shift is to evaluate not only whether an autonomous action works as intended, but also whether the environment that the action leaves behind remains inhabitable over time.

In conclusion, I would like to return to the same architectural space that I asked you to imagine at the beginning of this article. Robotic architecture could potentially help this space respond better to our changing needs, improve our wellbeing, and distribute resources more efficiently. Yet, as we have seen, once architecture can autonomously initiate reconfigurations, we also need to understand how we would continue to inhabit it, share it with others, and maintain some sense of control over it. As computing gains the capacity to physically reorganise inhabited environments, we need to understand not only how computational systems act, but also how occupants inhabit the conditions that their actions leave behind.

\bibliographystyle{ACM-Reference-Format}
\bibliography{main}

\end{document}